\documentclass[conference]{ieeeconf}
\IEEEoverridecommandlockouts
\usepackage{cite}
\usepackage{amsmath,amssymb,amsfonts}
\usepackage{algorithmic}
\usepackage{graphicx}
\usepackage{textcomp}
\usepackage{xcolor}
\usepackage{soul}
\usepackage{tikz}
\usepackage{booktabs}    
\usepackage{multirow}    
\usepackage{array}       
\usetikzlibrary{arrows.meta, positioning} 
\def\BibTeX{{\rm B\kern-.05em{\sc i\kern-.025em b}\kern-.08em
    T\kern-.1667em\lower.7ex\hbox{E}\kern-.125emX}}
    
\begin{document}

\title{Growing Perspectives: Modelling Embodied Perspective Taking and Inner Narrative Development Using Large Language Models}

\author{
Sabrina Patania$^{1,*}$, Luca Annese$^{1}$, Anna Lambiase$^{1}$, Anita Pellegrini$^{1}$,\\
Tom Foulsham$^{2}$, Azzurra Ruggeri$^{3}$, Silvia Rossi$^{4}$, Silvia Serino$^{1}$, Dimitri Ognibene$^{1,*}$
\thanks{*This work was supported by the Volkswagen Foundation under the funding programme “Open Up – New Research Spaces for the Humanities and Cultural Studies,” project “Developing an Artificial Social Childhood (ASC) to improve AI causal reasoning, information gathering and decision making,” reference 9E530.}
\thanks{$^{1}$Department of Psychology, University of Milan-Bicocca, Milan, Italy.
        {\tt\small \{sabrina.patania,dimitri.ognibene\}@unimib.it}}%
\thanks{$^{2}$Department of Psychology, University of Essex, Essex, UK.
        {\tt\small foulsham@essex.ac.uk}}%
\thanks{$^{3}$School of Social Sciences and Technology, Technical University of Munich, Munich, Germany.
        {\tt\small a.ruggeri@tum.de}}%
\thanks{$^{4}$Department of Electrical Engineering and Information Technology, University of Naples Federico II, Naples, Italy.
        {\tt\small silvia.rossi@unina.it}}%
}

\maketitle

\begin{abstract}
Language and embodied perspective taking are essential for human collaboration, yet few computational models address both simultaneously. This work investigates the PerspAct system \cite{patania2025perspact}, which integrates the ReAct (Reason and Act) paradigm with Large Language Models (LLMs) to simulate developmental stages of perspective taking, grounded in Selman’s theory \cite{selman1980growth}. Using an extended director task, we evaluate GPT’s ability to generate internal narratives aligned with specified developmental stages, and assess how these influence collaborative performance both qualitatively (action selection) and quantitatively (task efficiency). Results show that GPT reliably produces developmentally-consistent narratives before task execution but often shifts towards more advanced stages during interaction, suggesting that language exchanges help refine internal representations. Higher developmental stages generally enhance collaborative effectiveness, while earlier stages yield more variable outcomes in complex contexts. These findings highlight the potential of integrating embodied perspective taking and language in LLMs to better model developmental dynamics and stress the importance of evaluating internal speech during combined linguistic and embodied tasks.
    
\end{abstract}

\section{Introduction}

Perspective taking—the ability to represent a situation from an alternate viewpoint \cite{grice1975logic}—is fundamental for effective communication, joint action, and social coordination, not only in humans but increasingly in artificial agents designed to interact with people \cite{frith2005theory}. It encompasses distinct yet related processes: visual perspective taking (distinguishing what others can see versus how they see it \cite{flavell1981young}), and spatial perspective taking (representing relative spatial relationships between agents and objects through egocentric or allocentric reference frames). Both play a central role in collaborative problem solving, particularly when agents must reason about another's knowledge, perceptions, and intentions.

Recent research has explored whether Large Language Models (LLMs) can support such abilities by integrating explicit perspective cues to improve collaborative dialogue \cite{wilf2023think}. However, these approaches largely overlook insights from developmental psychology, where perspective taking is understood as an evolving capacity progressing through well-defined developmental stages. Selman’s theory of perspective taking \cite{selman1980growth} provides a comprehensive framework for modelling this progression: from early egocentric stages, where one fails to distinguish self from other, to more sophisticated forms of reciprocal and integrative perspective coordination.

Building on this theoretical foundation, we build upon our earlier PerspAct system \cite{patania2025perspact}, which originally integrated the ReAct (Reason and Act) paradigm with LLMs to support perspective taking in collaborative tasks. While the original PerspAct framework focused on generating and using internal narratives to guide reasoning and action, the present work extends this approach by explicitly incorporating developmental stages of perspective taking, inspired by Selman’s theory. In this version, PerspAct simulates child-like reasoning at varying stages of perspective development, allowing us to examine how developmental factors shape both internal representations and collaborative behaviour.

Specifically, we address two research questions: (i) can GPT reliably generate scene descriptions that reflect different developmental stages when prompted with adult-like allocentric inputs? and (ii) to what extent do developmentally-informed internal narratives influence PerspAct’s collaborative performance during task execution? To explore these questions, we adopt an extended director task designed to evaluate both the generation of perspective-aligned narratives and their impact on action selection and task efficiency across developmental conditions.

\section{Related Work}

Recent advancements in collaborative AI emphasize the importance of effective perspective taking to improve human-AI interactions. The foundational ReAct framework introduced by Yao et al. \cite{yao2023react} has enabled agents to reason and act efficiently in dynamic collaborative environments. Extending this framework, PerspAct \cite{patania2025perspact} integrates active vision and explicit perspective taking, demonstrating enhanced situated collaboration skills.

Developmental psychology literature addresses the gradual evolution of perspective taking abilities through various models. We choose to refer to Selman's model, where he establishes five stages of perspective taking \cite{selman1980growth, selman1971taking}:

\begin{itemize}
    \item \textbf{Egocentric (Stage 0, undifferentiated, before age 6)}: 
    The child physically distinguishes themselves from others but remains socially egocentric, assuming that other think as they do. 

    \item \textbf{Differentiated, subjective (Stage 1, ages 6–8)}: 
    Children acknowledge that others may hold perspectives different from their own, typically due to having access to different information, but still struggle to accurately judge others' viewpoints.

    \item \textbf{Self-reflective, reciprocal (Stage 2, ages 8–10)}: 
    Children can metaphorically "step into another person's shoes", recognizing and validating differing perspectives as appropriate given distinct situational contexts.

    \item \textbf{Mutual, third-person (Stage 3, ages 10–12)}: 
    Children begin viewing interactions from a more neutral, third-person perspective, simultaneously managing multiple viewpoints and integrating them into a shared understanding.

    \item \textbf{Societal (Stage 4, ages 12 and beyond)}: 
    Adolescents develop an understanding of how perspectives are embedded in broader societal frameworks, influenced by institutional roles, norms, and values. They recognize that an individual's perspective is shaped by their social context and position.
\end{itemize}

Recent cognitive science studies suggest perspective taking can be linked to linguistic competencies such as personal pronoun use, highlighting the importance of developmental linguistics in perspective taking \cite{ricard1999personal}. Nematzadeh et al. \cite{nematzadeh2018evaluating} evaluate computational models' capacity for inferring mental states within narrative contexts, underscoring the critical role of linguistic representation in perspective taking tasks. Moreover, Rashkin et al. \cite{rashkin2018modeling} emphasize the importance of empathetic and perspective-aware dialogue generation, highlighting how language models can be tailored to address complex social cognition scenarios.

A growing number of recent studies have also evaluated LLMs' implicit theory of mind capabilities across diverse tasks. For example, Chen et al. \cite{chen2024tombench} and Wang et al. \cite{wang2025rethinking} propose novel benchmarks for probing ToM reasoning in GPT-based models, while Zhou et al. \cite{zhou2023sotopia} investigate LLMs’ ability to manage social beliefs and predict others’ behaviour in interactive settings. These approaches, however, tend to focus on behavioural outcomes rather than grounding LLM reasoning in structured developmental frameworks.

Although wide research has been conducted on perspective taking, not many authors have adopted a developmental framework. In this work, we aim to bridge these developmental insights with contemporary AI by exploring how developmental theories can enhance collaborative capabilities of modern LLM frameworks.

\section{Method}

\begin{figure}[ht!]
    \centering
    \resizebox{0.9\linewidth}{!}{%
    \begin{tikzpicture}[auto,
        block/.style={rectangle, draw, align=center, rounded corners, font=\small, text width=4cm},
        expblock/.style={rectangle, draw, align=center, rounded corners, font=\small, fill=gray!10, text width=3cm},
        arrow/.style={-{Stealth}, thick}]
        
        \node[block] (scene) {Objective Scene Description};
        \node[block, below=0.6cm of scene] (gptgen) {GPT-4o: Child-like Narratives (Stages 0 and 1)};
        \node[block, below=0.6cm of gptgen] (evaluation) {Compliance Evaluation};
        \node[block, below=0.6cm of evaluation] (narratives) {Categorised Narratives};

        \node[expblock, below left=0.8cm and -1cm of narratives] (blind) {Blind};
        \node[expblock, below=0.8cm of narratives] (informed) {Informed};
        \node[expblock, below right=0.8cm and -1cm of narratives] (objinformed) {Objective-Informed};

        \node[block, below=1cm of informed] (finaleval) {Qualitative \& Quantitative Evaluation};

        \draw[arrow] (scene) -- (gptgen);
        \draw[arrow] (gptgen) -- (evaluation);
        \draw[arrow] (evaluation) -- (narratives);
        
        \draw[arrow] (narratives.west) -| (blind.north);
        \draw[arrow] (narratives) -- (informed);
        \draw[arrow] (scene.east) -| (objinformed);
        
        \draw[arrow] (blind.south) |- (finaleval.west);
        \draw[arrow] (informed.south) -- (finaleval.north);
        \draw[arrow] (objinformed.south) |- (finaleval.east);
        
    \end{tikzpicture}
    }
    \caption{Compact overview of the experimental setup and evaluation process.}
    \label{fig:method-compact}
\end{figure}
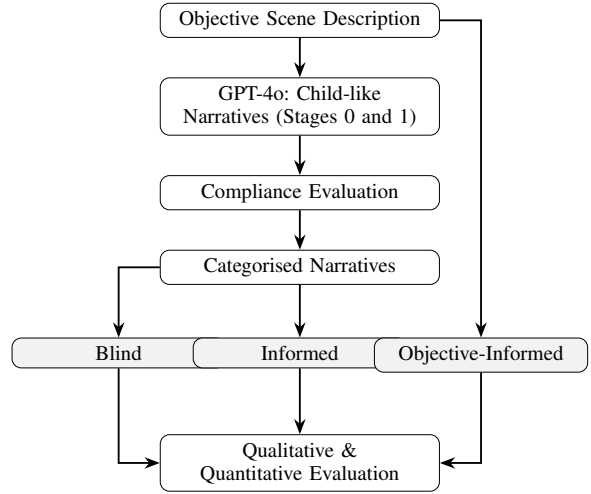

\subsection{Experimental Design}
We adapt the PerspAct experimental setup, employing scenarios drawn from an original dataset generated through GPT-4o. Scenarios are explicitly reformulated according to the developmental stages outlined by Selman and colleagues  \cite{selman1971taking, selman1975level, selman1980growth, elfers2008perspective}.

To pragmatically test the impact of developmental stages on LLM perspective taking, we simplify the five-stage theoretical framework focusing on two distinct stages representative of key developmental transitions:

\begin{itemize}
    \item \textbf{Egocentric (Corresponding to Selman's Stage 0)}: 
    Scenarios are described exclusively from the agent's viewpoint, ignoring others' viewpoints.
    \textit{Example prompt:} ``There's a blue tie and a red tie in the wardrobe, and a closed drawer at the bottom''.

    \item \textbf{Self-other, differentiated (Corresponding roughly to Selman's Stage 1)}: 
    Recognition that the other agent might not see everything from their position, thus partially adjusting the descriptions.
    \textit{Example prompt:} ``I clearly see the red tie and the closed drawer at the bottom, but from your side, you might not''.
\end{itemize}

Although Selman's complete model includes finer developmental distinctions, this simplified scheme was chosen to effectively explore core developmental impacts on collaborative LLM performance, balancing theoretical detail and experimental practicality.

\subsection{Procedure}
Each scenario was initially described in detail from an allocentric perspective (Objective Scene). GPT-4o was then prompted to rewrite these descriptions, emulating a child's perspective that corresponds to a specific Selman stage (Child-like Internal Narrative). For instance, for Stage 0, GPT was instructed to follow own action orientations, while at higher stages, prompts emphasised mutual awareness and sophisticated perspective integration. 

Additionally, to explore the impact of explicit developmental self-identification, GPT was prompted to adopt the cognitive characteristics typical of a child at each developmental stage, providing an additional layer of cognitive simulation. 


To operationally assess GPT-generated text according to developmental stages of perspective taking (Compliance Evaluation), and considering that, to the best of our knowledge, a standardised evaluation methodology for narrative compliance is yet to be established, we develop a structured evaluation protocol informed by existing theoretical insights, particularly those articulated in \cite{ricard1999personal} and \cite{selman2012social}. While this integrative approach reflects a pragmatic effort to leverage available theoretical frameworks, we recognise it as a foundational step that future research may refine and expand to fully encompass the complexity inherent in developmental perspective taking.

The following operational criteria were defined for evaluating GPT-generated texts:

\begin{itemize}
    \item \textbf{Pronouns (Pron)}: Analysis of pronoun use to distinguish between self and others, crucially linked to early perspective taking skills \cite{ricard1999personal}.
    \item \textbf{Actors taken into account (Actors)}: Identification of individuals explicitly acknowledged in the text as possessing distinct perspectives \cite{selman2012social}.
    \item \textbf{Self perspective articulation (SelfArt)}: Degree of explicit description regarding the agent’s own potential thoughts, feelings, or orientations to action.
    \item \textbf{Director’s perspective articulation (DirArt)}: Explicit description of the director’s potential thoughts, feelings, or orientations to action.
    \item \textbf{Awareness of limits from self perspective (SelfLim)}: Recognition of limitations in the agent’s own viewpoint (acknowledging that the agent cannot see or know everything).
    \item \textbf{Awareness of limits from director’s perspective (DirLim)}: Recognition that the director’s viewpoint is similarly constrained.
\end{itemize}

These criteria were evaluated according to two simplified developmental stages (adapted from Selman’s full five-stage model):

\subsection*{Stage 0: Egocentric (undifferentiated, before age 6)}
Texts are primarily egocentric with limited perspective differentiation:
\begin{itemize}
    \item \textit{Pron}: Predominantly first-person pronouns (\textit{I, me}), minimal differentiation from others.
    \item \textit{Actors}: Self-focused; others’ perspectives are generally conflated with self.
    \item \textit{SelfArt}: Limited and often inaccurate; self-thoughts projected onto others.
    \item \textit{DirArt}: Minimal or incorrect; assumes others share identical perspectives.
    \item \textit{SelfLim}: Absent; presumption of complete self-knowledge.
    \item \textit{DirLim}: Absent; fails to recognize others’ differing viewpoints.
\end{itemize}

\subsection*{Stage 1: Differentiated, subjective (ages 6–8)}
Texts reflect initial differentiation between self and others:
\begin{itemize}
    \item \textit{Pron}: Use of \textit{I} and \textit{they}; emerging distinction between self and others.
    \item \textit{Actors}: Acknowledges others exist but emphasizes self-perspective.
    \item \textit{SelfArt}: Recognizes own perspective explicitly.
    \item \textit{DirArt}: Attempts but incomplete; differences attributed primarily to informational gaps.
    \item \textit{SelfLim}: Emerging recognition of viewpoint limitations.
    \item \textit{DirLim}: Limited recognition; incomplete understanding of another’s viewpoint.
\end{itemize}

This structured evaluation protocol enabled the operational assessment of GPT-generated texts, systematically linking developmental perspective taking theory to quantifiable textual characteristics.

Each generated scenario (Categorised Narratives) was used as input to the PerspAct framework (ReAct-based matcher agent). Task performance was evaluated based on accuracy and efficiency in identifying objects from collaborative tasks requiring perspective taking.

A complete overview of the procedure's flow is illustrated in Fig. \ref{fig:method-compact}.

\subsection{Hypotheses}
Our experimental hypotheses are twofold:
\begin{enumerate}
    \item GPT can reliably and consistently produce scenario descriptions aligned with distinct developmental stages, demonstrating internal coherence relative to developmental theory.
    \item Perspective taking accuracy and collaborative performance improve progressively from Stage 0 (Egocentric) to higher stages.
\end{enumerate}

These experiments aim to provide insights into the cognitive plausibility and performance benefits of developmentally-informed perspective taking prompts for AI systems.

\section{Experiments}


We conducted two experiments based on extension of the director task designed to address our primary research questions. Both experiments utilised a set of seven perspective taking scenarios—five classic and two ecological—where the task of the agent, as described in the original PerspAct setup \cite{patania2025perspact}, involves identifying objects in collaboration with another actor (the director), whose visibility and perspective differ from that of the agent. While classic scenarios feature static environments and limited action space, the ecological scenarios introduce more realistic interaction dynamics: the matcher is allowed to move within the environment, which is itself more complex and less constrained. This distinction enables us to assess the generalisability and robustness of developmental stage-driven reasoning across varying levels of perceptual and environmental complexity.

\subsection{Internal Narrative Generation}

In the first experiment, GPT-4o was tasked with generating child-like narrative descriptions corresponding to two simplified developmental stages (Egocentric, Differentiated, Reciprocal) from a given objective scenario description. To evaluate the developmental compliance of the GPT-generated narratives, we employed our operational criteria: Pronouns (Pron), Actors taken into account (Actors), Self perspective articulation (SelfArt), Director’s perspective articulation (DirArt), Self perspective limits awareness (SelfLim), and Director’s perspective limits awareness (DirLim). GPT-4o is also used to evaluate these descriptions, rating their alignment with the defined developmental criteria.

Examples of narrative outputs for each stage are as follows:
\begin{itemize}
    \item \textbf{Stage 0 (Egocentric)}: \textit{“I see a blue\_tie hanging from the wardrobe, just to the left. There is a shirt hanging next to it, next to the blue\_tie. On the shelf, I also see a red\_tie that is folded. And at the bottom, there is a closed drawer. That's all I see.”}

    \item \textbf{Stage 1 (Differentiated)}: \textit{"I see a wardrobe with a blue tie hanging on the left and a red tie folded on a shelf. There's also a closed drawer at the bottom, but I don’t know what’s inside. Next to the blue tie, there’s a shirt hanging. The other person sees the wardrobe too, and they can also see the blue tie and the shirt. But they can’t see the drawer, so they don’t know what’s inside. I can see everything, so I know more about what’s there."}

\end{itemize}

In our evaluation, GPT-4o correctly categorised the first two scenarios according to their respective developmental stages.  

\subsection{Collaborative Task Performance}
In the second experiment, we evaluate how effectively PerspAct completed collaborative tasks based on the GPT-generated narratives from the previous experiment. Each scenario is evaluated 5 times, for the stochastic nature of the task, to ensure consistency. Three sub-experiments were conducted:

\paragraph{Blind Experiment (without developmental stage information)} We provided GPT-o3-mini with the child-like narrative descriptions without specifying the developmental stage. This allowed us to assess the model's inherent sensitivity to narrative descriptions of different developmental complexities.

\paragraph{Informed Experiment (with developmental stage information)} In the second sub-experiment, we explicitly informed GPT-o3-mini about the developmental stage corresponding to each narrative. This aimed to examine whether an explicit developmental context could further improve collaborative performance.

\paragraph{Objective-Informed Experiment (allocentric description with developmental stage information)} In the third sub-experiment, we provide GPT-o3-mini with an objective, allocentric description of the scene (the one used for the Internal Narrative generation), along with explicit information about the developmental stage it is meant to simulate. This condition enables us to assess the model's ability to generate actions based on a more neutral representation of the environment, while still integrating the intended developmental framing. By decoupling the child-like narrative from the initial input, this experiment serves to evaluate how developmental cues alone influence the model’s perspective taking and task performance.

To evaluate performance, we record the number of steps taken to successfully complete the collaborative tasks for each developmental narrative, analysing efficiency gains associated with advanced perspective taking stages.

\begin{table}[ht]
\centering
\caption{Average percentage of task failures across experimental conditions (classic scenarios) and developmental stages.}
\label{tab:failure-rates}
\begin{tabular}{lcc}
\toprule
\textbf{Condition} & \textbf{Stage 0 (\% failures)} & \textbf{Stage 1 (\% failures)} \\
\midrule
Blind & 0\% & 0\% \\
Informed & 0\% & 25\% \\
Objective-Informed  & 0\% & 0\% \\
\bottomrule
\end{tabular}
\end{table}
\vspace{-10pt}
\begin{table}[ht]
\centering
\caption{Average number of steps required to complete the task (classic condition) across experimental conditions and developmental stages.}
\label{tab:steps-classic}
\begin{tabular}{lcc}
\toprule
\textbf{Condition} & \textbf{Stage 0 (avg. steps)} & \textbf{Stage 1 (avg. steps)} \\
\midrule
Blind & 1.75 & 1.75 \\
Informed & 1.75 & 0.94 \\
Objective-Informed & 1.75 & 1.75 \\
\bottomrule
\end{tabular}
\end{table}

\begin{table}[ht]
\centering
\caption{Average percentage of task failures in the ecological condition across experimental configurations and developmental stages.}
\label{tab:failure-ecological}
\begin{tabular}{lcc}
\toprule
\textbf{Condition} & \textbf{Stage 0 (\% failures)} & \textbf{Stage 1 (\% failures)} \\
\midrule
Blind & 30\% & 0\% \\
Informed & 40\% & 0\% \\
Objective-Informed & 20\% & 0\% \\
\bottomrule
\end{tabular}
\end{table}

\begin{table}[ht]
\centering
\caption{Weighted average number of steps in the ecological condition, combining both conditions  weighted by task success rate.}
\label{tab:steps-ecological-weighted}
\begin{tabular}{lcc}
\toprule
\textbf{Condition} & \textbf{Stage 0 (weighted)} & \textbf{Stage 1 (weighted)} \\
\midrule
Blind & 6.29 & 6.80 \\
Informed & 7.20 & 6.60 \\
Objective-Informed & 6.75 & 6.60 \\
\bottomrule
\end{tabular}
\end{table}
\vspace{-10pt}
Across both the classic and ecological conditions, results reveal consistent patterns. In the classic condition, task failures were negligible at both developmental stages, with a slight performance drop in the Informed condition at Stage 1. Step counts remained relatively stable, suggesting uniform task complexity across conditions.

In contrast, the ecological condition introduced more variability. At Stage 0, failure rates were higher across all configurations, especially in the Informed condition, indicating greater difficulty in generalising simpler developmental reasoning to dynamic settings. However, Stage 1 performance remained robust across all experiments.

When adjusting the step count by task success rates and combining across trials, Stage 1 showed consistent efficiency across all conditions. Stage 0, by contrast, showed greater variation: the Informed condition required more steps, while the Blind setup proved most efficient. 

\subsection{Internal Narrative Evaluation}

To assess GPT-4o's ability to generate developmentally appropriate perspective taking narratives, we conduct an evaluation experiment focusing on two key phases: (i) the Internal Narrative generated from the previous phase from the Objective Scene Description prior to any interaction with the director (Pre-task Specification), and (ii) the reasoning expressed during the interaction, once the director’s request had been introduced (Post-task Execution). For both phases, we aim to verify the consistency of the generated narratives with the developmental stages of Egocentric and Differentiated perspective taking, using the operational criteria introduced earlier.

In the Pre-task Specification phase, GPT-4o is prompted to simulate a child at a specific developmental stage and describe the scene from child’s point of view. This produces Internal Narratives that were evaluated based on the six defined criteria (Pronouns, Actors, SelfArt, DirArt, SelfLim, DirLim) to assess their developmental alignment.

In the Post-task Execution phase, we evaluate the reasoning steps generated during the task resolution under three experimental conditions:

\begin{itemize}
    \item Blind, in which the system had access only to the child-like narrative;
    \item Informed, where the system received both the narrative and explicit information about the developmental stage;
    \item Objective-Informed, in which the system was given the allocentric description of the scene along with the stage specification.
\end{itemize}

The average scores across conditions and developmental stages are reported in Table~\ref{tab:evaluation}. These scores (0 or 1) reflect whether the GPT-genereted outputs are conformed to the establish criteria for Egocentric and Differentiated stages. 

The results show that in the Pre-task Specification phase, the Internal Narratives generated for both Egocentric and Differentiated stages fully fit the corresponding criteria. This indicates that GPT-4o is capable of simulating developmentally coherent internal perspectives prior to interaction.

However, in the Post-task Execution phase, a marked difference emerged. Internal narratives corresponding to the Differentiated stage consistently satisfied the evaluation criteria across all conditions. In contrast, outputs associated with the Egocentric stage failed to meet the required criteria, suggesting that once the interaction with the director begins, GPT-4o tends to shift toward more advanced reasoning, incorporating the other agent’s perspective regardless of the instructed developmental stage.

This behavioral shift may be attributed not to an inherent limitation in the model's generative capacity, but to the limitations of applying criteria designed for static narrative analysis to a dynamic, interactional context. The presence of a linguistic request appears to act as a trigger for more socially attuned representations, and future work should consider developing adapted criteria to distinguish developmentally plausible responses in dyadic interactions.

\begin{table}[ht]
    \centering
    \setlength{\tabcolsep}{4pt}
    \begin{tabular}{l *{4}{c}}
        \toprule
         & \multicolumn{2}{c}{Pre-task specification} & \multicolumn{2}{c}{Post-task specification} \\
         & Egocentric & Differentiated & Egocentric & Differentiated \\
        \midrule
        Egocentric & 0 & 0 & 0 & 1 \\
        Differentiated & 0 & 1 & 0 & 1 \\
        \bottomrule
    \end{tabular}
    \vspace{1ex} 
    \caption{Confusion matrices of Narratives Evaluation}  \label{tab:evaluation}
\end{table}\vspace{-10pt}

\section{Discussion}
Our findings suggest that GPT generates narratives that mimic patterns associated with developmental stages, effectively reflecting the increasing complexity of perspective taking abilities. Notably, our results clearly reflect the impact of developmental stage on the collaborative efficiency of PerspAct. Specifically, the Egocentric Stage (Stage 0) narratives resulted in consistent task failures, underscoring the fundamental limitations in collaborative comprehension inherent at this stage. Conversely, the Differentiated (Stage 1) narratives yielded significant improvements in both classic and ecological experiments.

These findings align closely with developmental psychology literature, emphasizing enhanced cognitive and linguistic competencies associated with advanced perspective taking stages. The progression observed in our study underscores the importance of integrating developmental theory into AI models, suggesting that explicitly considering developmental stages can substantially enhance model accuracy and efficiency. Moreover, it supports the capacity of GPT to simulate the initial child’s point of view in a consistent and structured way, as evidenced by the narrative evaluations. 

Nevertheless, our study includes certain limitations. Primarily, we focus only on the first two stages of  Selman, leaving  the exploration of more nuanced shifts in perspective taking for future work. Additionally, our evaluation approach, while systematic and grounded in theory, necessarily reflects operational choices made in the absence of a universally accepted methodology for evaluating developmentally appropriate narrative generation. This highlights the need for more robust and fine-grained tools to analyse linguistic perspective taking in AI.

Another important dimension of our findings relates to the ability of the LLM to correctly generate and classify the child-like internal narrative. More specifically, we must distinguish between evaluating the narrative GPT produces as its initial description of the scene, and the reasoning that emerges after the director’s explicit request. Interestingly, the initial linguistic perception of the model of the scene often aligns well with the requested developmental stage—even in Stage 0. However, after the introduction of the director's request, the model shifts to a Stage 1 classification, reflecting an increased responsiveness to linguistic cues that suggest an increased awareness of the director's presence. 

This shift does not necessarily indicate that the model is incapable of maintaining early-stage thinking. Rather, it may suggest a limitation in our classification criteria, which were primarily designed to assess monologic descriptions. These may not be optimally suited for interpreting perspective taking within an interactional, dyadic context where pragmatic cues and expectations are more prominent. Future work could aim to develop complementary criteria specifically tailored to assess responses that unfold in dialogue or interaction.

The observed asymmetry also highlights a deeper theoretical distinction: that between embodied and linguistic forms of perspective taking \cite{}. The former may arise when a child (or agent) perceives others in a spatial and sensory manner; the latter becomes more salient when social demands are introduced through language. Addressing someone directly, particularly in the form of a question, imposes a pragmatic demand that naturally invites a naive form of perspective taking, or something easily classified as such. In this sense, linguistic input acts not merely as information but as a trigger for representational alignment. This dynamic is particularly relevant to LLMs, which rely exclusively on textual information: their representation of the "other" is always mediated by language, and therefore may favour the emergence of a social-linguistic model of the other.

It is plausible to hypothesise the existence of at least two partially independent systems for representing others: one grounded in perceptual and motor schemas, and another shaped by linguistic interaction. The latter, inherently social and structured by dialogue, may be more readily accessible to LLMs. If one can speak, it is as if the other is always already present in the linguistic landscape.

Another factor that may shape the model’s behavior is its implicit bias toward information-seeking strategies. The availability of a mechanism to ask questions ("ask" is a possible action to perform) may itself introduces a subtle expectation that the director knows the location of the object, reinforcing the tendency to treat the director as an intentional agent with independent goals and knowledge. While useful in many contexts, this could lead to inflated perspective taking at lower developmental stages. Exploring how the affordance of asking questions influences reasoning and role assignment in GPT could be an important avenue for future work.

Ultimately, the nuanced behaviors observed in GPT open a space for investigating how perspective taking unfolds not just as a static classification of outputs, but as a dynamic cognitive style influenced by interactional structure, narrative framing, and the affordances of language itself. By viewing large language models through a developmental lens, we gain not only a clearer picture of their limitations, but also a powerful new framework for studying the emergence of social cognition in artificial agents. 

\bibliographystyle{IEEEtran}
\bibliography{mybibliography}

\end{document}